\pdfoutput=1

\documentclass[11pt]{article}

\usepackage{ACL2023}

\usepackage{times}
\usepackage{latexsym}

\usepackage[T1]{fontenc}

\usepackage[utf8]{inputenc}

\usepackage{microtype}
\definecolor{burgundy}{rgb}{0.5, 0.0, 0.13}
\definecolor{brown(web)}{rgb}{0.65, 0.16, 0.16}
\definecolor{carmine}{rgb}{0.59, 0.0, 0.09}
\definecolor{forestgreen(web)}{rgb}{0.13, 0.55, 0.13}
\definecolor{emerald}{rgb}{0.31, 0.78, 0.47}
\definecolor{forestgreen(traditional)}{rgb}{0.0, 0.27, 0.13}

\usepackage{inconsolata}
\usepackage{graphicx}
\usepackage{multirow}
\usepackage{algorithm2e}
\RestyleAlgo{ruled}
\usepackage{longtable}
\usepackage{empheq}
\usepackage{amsmath}
\usepackage{geometry}
\SetKwComment{Comment}{}{}
 
\usepackage{xcolor}
\title{SQLPrompt: In-Context Text-to-SQL with Minimal Labeled Data}

\author{
Ruoxi Sun$^1$, Sercan \"{O}. Arik$^1$, Rajarishi Sinha$^1$,  Hootan Nakhost$^1$, \\
\textbf{Hanjun Dai$^2$, Pengcheng Yin$^2$, Tomas Pfister$^1$ }  \\
  $^1$ Cloud AI Research Team \\
  $^2$ Google DeepMind \\
   \texttt{\{ruoxis, soarik, sinharaj, hootan, hadai, pcyin, tpfister  \}@google.com} \\
}

\begin{document}
\maketitle
\begin{abstract}
Text-to-SQL aims to automate the process of generating SQL queries on a database from natural language text. In this work, we propose "SQLPrompt", tailored to improve the few-shot prompting capabilities of Text-to-SQL for Large Language Models (LLMs). Our methods include innovative prompt design,  execution-based consistency decoding strategy which selects the SQL with the most consistent execution outcome among other SQL proposals, and a method that aims to improve performance by diversifying the SQL proposals during consistency selection with different prompt designs ("MixPrompt") and foundation models ("MixLLMs"). We show that \emph{SQLPrompt} outperforms previous approaches for in-context learning with few labeled data by a large margin, closing the gap with finetuning state-of-the-art with thousands of labeled data. 
\end{abstract}

\section{Introduction}


Text-to-SQL enables natural language interfaces for databases via SQL query generation. It is crucial for enhancing database accessibility without SQL expertise, and enabling the development of conversational agents with data analysis ability.


Language models (LM) have shown to be promising for Text-to-SQL. Notable previous work on finetuning, including PICARD \cite{scholak2021picard}, UnifiedSKG \citep{xie2022unifiedskg}, and RESDSQL-3B + NatSQL\citep{li2023decoupling}, achieve impressive results by leveraging customized SQL-specific syntax knowledge and training on a large number of (text, SQL) paired data samples.   
Recently, large language models (LLMs) such as GPT-3 \citep{brown2020language}, PaLM \citep{chowdhery2022palm}, and ChatGPT\footnote{ \url{https://chat.openai.com/chat}.}\cite{stiennon2020learning} have demonstrated promising few-shot abilities via  prompting\citep{wei2022emergent}. By only providing a few demonstrations in the prompt, LLMs are able to follow the demonstrations and generate reasonable answers (``in-context learning'').  For Text-to-SQL, few-shot prompting is beneficial as it does not require expensive training, lowers adaptation data requirements, reduces out-of-distribution issues (e.g. for unseen phrases), and reduces the risk of over-fitting and poor generalization. \\In this paper, we introduce ``SQLPrompt'', a few-shot prompting approach for Text-to-SQL. The approach is comprised of execution-based consistency decoding and execution error filtering, ``MixPrompt'' and ``MixLLMs'' to enhance diversity of SQL proposals. In few-shot prompting, self-consistency decoding \citep{wang2022self}, which samples a diverse set of reasoning paths and selects the most consistent answer, has shown remarkable performance improvements across different tasks. Here we propose a novel variant of consistency decoding for Text-to-SQL, ``execution-based consistency decoding and execution error filtering''. The proposed decoding is tailored to be SQL-execution specific -- we marginalize over SQLs, and conduct majority vote on execution outcome, whereas the original self consistency approach marginalizes over chain-of-thought thinking path, and conduct majority vote on the final answer. \\
Furthermore, the improvement brought by self-consistency' decoding using the same prompt and same LLMs saturate beyond a certain number of samples. Because the diversity of SQL candidates are limited with the same prompt and same LLM. 
Here we propose employing diverse prompt designs ("MixPrompt") and LLMs ("MixLLMs") to obtain more diverse LLM outputs. "MixPrompt" applies different prompt designs, which changes the interface of query and LLMs, leading to more diverse LLM's outputed SQLs. "MixLLMs" holds the assumption that different LLMs pretrained on diverse set of knowledge, can yield different outcomes. 
Finally, the answer is the consistent answer across different prompt designs and LLMs. 
\section{Methods}

\subsection{Problem setup for Text-to-SQL}
Let $q$ be natural language query and $D_q$ be the database information. Text-to-SQL task is to convert query $q$ into SQL. 
The database $D_q = \{S, K_p, K_f\}$ includes database schema $S$, primary keys $K_p$, and foreign keys $K_f$.
$S$ usually contains multiple tables $T_t$: $S = \{T_1, T_2, ... T_t ... \}$. Each table $T_t$ has table name $N_t$, column names $c_j$ and column data types 
$t_j$: $T_t =\{N_t, (c_{t1}, t_{t1}), (c_{t2},t_{t2}), (c_{tj},t_{tj}) ... \}$)
Primary keys $K_p$ uniquely identifying rows of each table, and foreign keys $K_f$ join multiple tables.

\begin{figure*}[h]
\centering
\includegraphics[trim=0 550 100 60,clip,width= .85\textwidth]{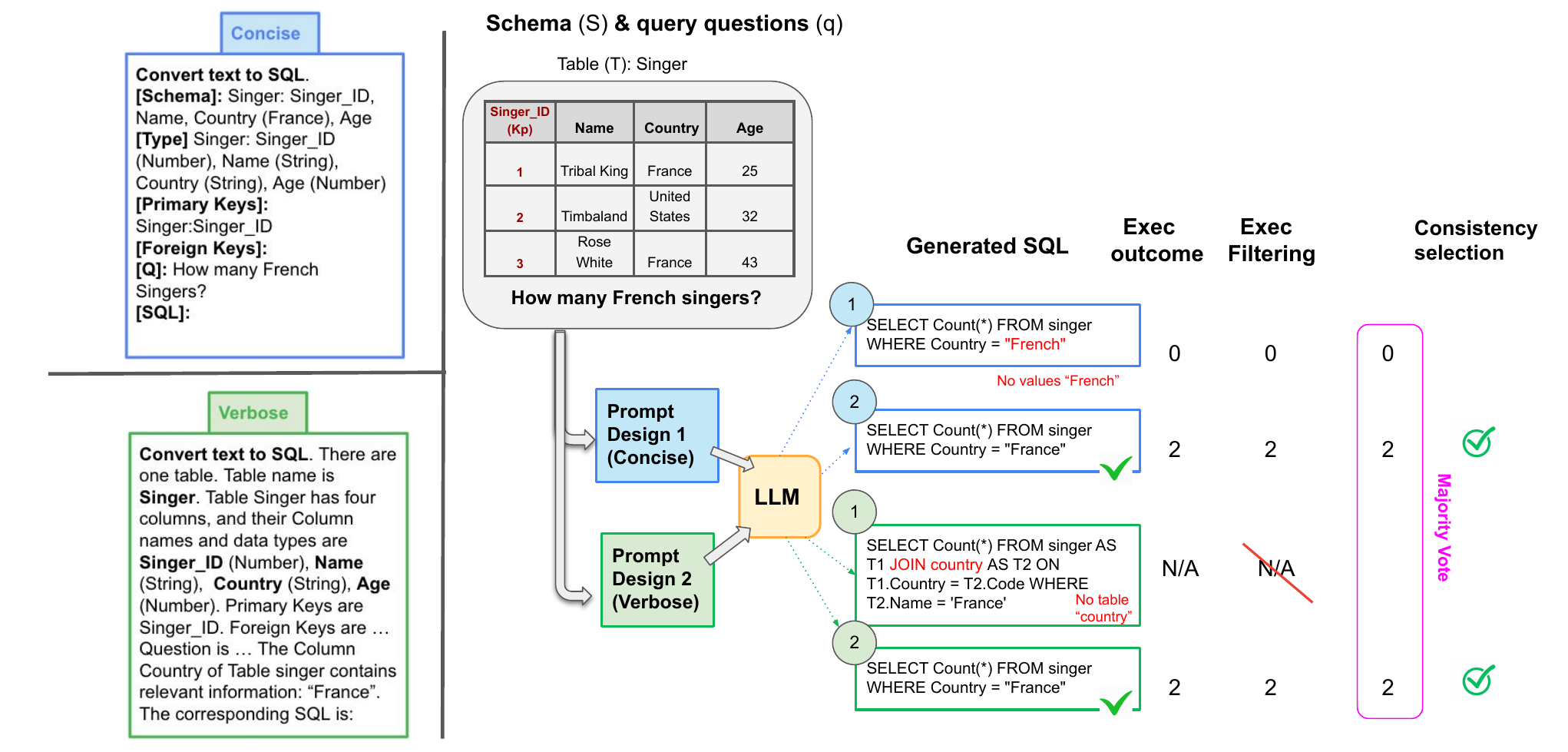}
\caption{
\textbf{SQLPrompt Overview}: \textbf{(Left)} \textbf{Prompt Design}: Concise prompt design (up) and Verbose prompt design (down).  \textbf{(Right)} \text{MixPrompt} in \emph{SQLPrompt} generates multiple prompts using database and query question, to query LLMs. For each query, LLMs are sampled twice, and two SQLs are generated and executed on the database with errors filtered out. The execution outcomes of both prompt designs are combined to select the most consistent SQL. Without MixPrompt, the true answer cannot be selected with only one prompt (blue) due to a tie situation. 
}\label{fig:SQLprompt Overview}
\vskip -0.5cm
\end{figure*}


\subsection{Prompt design: database schema and content and primary/foreign keys} \label{sec:design}
The prompt should include all necessary information needed for humans to generate SQL. The prompt is comprised of database schema, primary and foreign keys, and the database content. We reflect database content (entry values) as proposed in ~\citep{lin2020bridge,wang-etal-2020-rat}, where only values that are relevant to the question are included (refer to Appendix \ref{sec:promptdesign-explained} for more discussion). Furthermore, we present the above information in different formats, with the goal of making them different from each other to encourage diverse outputs. 

\textbf{Concise prompts}
we linearize information in a table as 
{\it``Table1 name: column name 1, column name 2 (relevant database content)} | {\it Table2 name: column1 ...''} (Figure~\ref{fig:SQLprompt Overview}, \textit{Concise}. Full example in Appendix~\ref{sec:concise}). 
This way describes table structure clearly, but can be less straightforward for LLMs to understand the syntax. \textbf{Verbose prompts}
we describe databases with human understandable words and emphasize on the information LLMs need to know: e.g. {\it ``Table ``CarNames'' contains three columns. The column names and their types are : MakeID (number), Model (string) ..''; ``Foreign keys are .. Use foreign keys to join Tables''}. See Appendix~\ref{sec:verbose} for an example.

\subsection{Refinement based on execution-based consistency with MixPrompt and MixLLMs} \label{sec:mixprompt+consistency}


We use few-shot prompting with execution-based consistency decoding and error filtering. We apply various prompt designs ("\textbf{\emph{MixPrompt}}") to encourage LLMs to generate diverse SQL outputs 
Suppose $F=\{f_1, f_2, ...\} $ is a collection of prompt design functions, e.g. $f_1$ is verbose, $f_2$ is concise.\ 
When we fix the LLMs, we have \textbf{\emph{MixPrompt}} with the following prediction objectives:  
\begin{align}
   p(sql | \text{LLM}, q) =  \sum_f p( sql |\text{LLM}, f, q) p(f),
\end{align}
where $p(f)$ is mixing coefficient. We evenly mix the prompts, hence, $p(f)=1/nF$, where $nF$ is the number of design functions. $p( sql |\text{LLM}, f, q)$ is the sampling probability of generating $sql$.

\begin{algorithm}[hbt!]
\small
\caption{Refinement based on execution and consistency with MixPrompt}\label{alg:mixprompt-alg}
\KwData{Require: Query questions $Q_{test}$; Database $D_{test}$; 
Prompt design function collections $F$; B is consistency sample numbers} 
\KwResult{SQL output of test set: SQL$_{test}$}
\While{$q$  in $Q_{test}$}{
$D_q  \gets D_{test}[q]$\;
\While{$f$ in $F$}{
$\text{Prompt}_{qf}  \gets f(q_i, D_q)$ \Comment*[r]{eq~\eqref{eq:llm-prompt}}  
M = []\;
O = []\;
\While{$b$ in $B$}{
   $ sql_{qfb} \underset{\tiny{i.i.d}}{\sim} \text{LLM}(\text{Prompt}_{qf})$  \Comment*[r]{eq~\eqref{eq:llm-sample}}    
   $O_{qfb} = \emph{Exec}(sql_{qfb}, D_q)$ \;
   
  \If{"error" NOT in $O_q$}{
    $M \gets  sql_{qfb} $\;
    $O \gets  O_{qfb} $\;
  }
}
}
$sql_{select} = \{sql_q: O_q = Majority (O), q \in M \}$  \Comment*[r]{eq~\eqref{eq:majorityvote}}  
$\text{SQL} _\text{test}  \gets sql_{select}$
}
\end{algorithm}

\emph{MixPromt} is overviewed in Fig~\ref{fig:SQLprompt Overview}. For each design function $f$, we generate prompts using database $D_q$ and the query $q$. 
The trained LLMs specify the distribution $\ell: q \rightarrow \emph{sql}$, where we can draw sample from: 
\begin{align}
&\text{Prompt}_{qf}  = f (q, D_q)  \label{eq:llm-prompt} \\
&sql_{qf} \underset{\tiny{i.i.d}}{\sim} 
\text{LLM}(\text{Prompt}_{qf})  \label{eq:llm-sample}
\end{align}
We sample $B$ times from the LLM with the same prompt $\text{Prompt}_q$ to get SQL collections by Eq~\refeq{eq:llm-sample}:
\begin{align}
M_{qf} = \{sql_{qf}^1, ... sql_{qf}^b \}_B   \label{eq:sqls}
\end{align}
We then execute the generated SQLs using the engine $Exec$ (i.e. \emph{sqlite3}), which yields the outputs $O$ as the execution result of SQL on the provided database. \label{eq:sample} 
\begin{align}
\tiny O_{qf}=\{O^b_{qf}: O^b_{qf} = \emph{Exec}(sql_{qf}^b, D_q),   sql^b_q \in M_{qf}\}  \label{eq:exec} 
\end{align}
We further exclude outputs $O_{qf}$ that yield errors and only keep the valid output, therefore, obtain final (SQL, outcome) pairs for prompt design $f$:  $R_{qf} = (M_{qf}, O_{qf})  =  \{(M^b_{qf}, O^b_{qf}): O^b_{qf}  \neq \text{errors} \}$. 
We repeat the above process for each prompt design function $f$ and generate $R_q=\{R_{q1}, ... R_{qf}, ..\}_{nF}$, by concatenating all the results across multiple designs and obtain: 
\begin{align}
M_q &= [ M_{q1}, ..., M_{qf}... , M_{nF}] \\
Q_q &= [ O_{q1}, ..., O_{qf} ..., O_{nF} ]
\end{align}


Following self-consistency, we select the SQL outputs that give the execution outcome consisted with the majority of the execution outcomes $O_q$ generated by all $M_q$. 
\begin{align}
sql_{select} = \{ &sql_q^k: O_q^k = Majority(O_{q}) \\
&O_q^k \in Q_q, sql_q^k \in M_q \}, \label{eq:majorityvote}
\end{align}
where $k$ is the index across multiple prompt design and consistency repeats. The overall process is described in Algorithm \ref{alg:mixprompt-alg}. \\
\textbf{MixLLMs} With the goal of increasing diversity of the SQL proposals in consistency decoding, we further expand our method to not only use one LLM, but rather a mixture of LLMs. The consistency samples include resource from different prompt designs and different LLMs: 
\begin{align}
   \tiny p(sql|q) =  \sum_{\text{LLM}} \sum_f p(sql|LLM, f, q) p(f)p(LLM)
\end{align}
Similar to the combination idea in \emph{MixPrompt}, \emph{"MixLLM"} combines outputs across multiple LLMs, in addition to multiple prompt designs.  

We note that our method differs from generic Mixture of Expert (MoE) \cite{chen2022towards,zhou2022mixture} approaches as we instantiate MoE in few-shot prompting setup, where experts are various prompt designs. Moreover, rather than simple averaging, we combine results based on execution outcomes.


\begin{table*}[ht]
\vskip -5mm
\centering
\resizebox{0.56\textwidth}{!}{%
\begin{tabular}{llcc }
\hline
 \multirow{2}{*}{ } & \multirow{2}{*}{\textbf{Methods}} & \multicolumn{2}{l}{\textbf{SPIDER}} \\ 
                                  &  & \textbf{EX} & \textbf{TS} \\ \hline
 \multirow{3}{*}{\textbf{Fine-tuning}} & T5-3B + PICARD                      &  79.3  &  69.4      \\
& RASAT + PICARD                      &  80.5  & 70.3       \\
& RESDSQL-3B + NatSQL              &  \textbf{84.1}  &  \textbf{73.5}      \\ \hline
 \multirow{8}{*}{\textbf{In-context learning}} 
& GPT-3 ada (0-shot)                  &  2.3 & 0.3      \\
& GPT-3 babbage (0-shot)                 & 5.7  & 3.9      \\
& GPT-3 curie (0-shot)                 & 12.6  & 8.3      \\
& GPT-3 davinci (0-shot)                  & 26.3  & 21.7      \\
& Codex cushman (0-shot)                  & 63.7  & 53.0      \\
& Codex davinci (0-shot)                   & 67.0  & 55.1      \\
& ChatGPT (0-shot)                             &    70.1  &   60.1         \\ 
& SQLPrompt (0-shot)                 & \textbf{76.6}  & \textbf{68.0}  \\
& SQLPrompt (4-shot)                 & \textbf{77.1}  & \textbf{68.6}      \\ \hline
\end{tabular}%
}
\vskip -1mm
\caption{\small Performance on the Spider Dev set, measured in execution accuracy (EX) and test-suite accuracy (TS). GPT3 and CodeX results are from \citep{rajkumar2022evaluating} and ChatGPT results are from \citep{liu2023comprehensive}. }
\vskip -5mm
\label{tab:sota}
\end{table*}
 
\begin{table}[h]
\centering
\caption{\small Ablation study on prompt design approaches in 0-shot setting. MixPrompt improves concise or verbose prompt design approaches with different LLMs. We only mark TS Acc. changes, not EX, because TS is more accurate evaluation.}
\vskip -2mm
\resizebox{0.5\textwidth}{!}{
\begin{tabular}{lcccccc}
\hline

\multirow{2}{*}{Models} &  \multicolumn{2}{c}{\textbf{Concise}} & \multicolumn{2}{c}{\textbf{Verbose}} & \multicolumn{2}{c}{\textbf{MixPrompt}}  \\ 
 & EX & TS & EX & TS & EX & TS \\ 
\hline
PaLM FLAN 62B qntz  &  67.7  &  61.3 & 70.8 & 62.9 & 70.5 & 63.2 $({\color{teal} \uparrow 0.3})$ \\
PaLM FLAN 540B qntz  & 72.3 &  64.1 & 71.6   & 61.3 &    74.0 & 65.5 $({\color{teal} \uparrow 1.4})$ \\ 
\hline

\end{tabular}
}
\label{tab:zeroshots}
\vskip -4mm
\end{table}

\begin{table}[h]
\caption{\small Ablation Study: Few-shots}
\vskip -3mm
\resizebox{0.5\textwidth}{!}{%
\begin{tabular}{lcccccc}
\hline

\multirow{2}{*}{Models}  & \multicolumn{2}{c}{\textbf{Concise}} & \multicolumn{2}{c}{\textbf{Verbose}} & \multicolumn{2}{c}{\textbf{MixPrompt}}  \\ 
  & EX & TS & EX & TS & EX & TS \\ 
\hline
PaLM FLAN 62B qntz     &  65.9  & 59.6   & 71.8  & 63.8  & 74.7  &  66.6 $({\color{teal} \uparrow 2.8})$\\
PaLM FLAN 540B qntz     &  71.2 &  63.2  &  70.7  & 61.1  & 74.7     &  65.2 $({\color{teal} \uparrow 2.0})$ \\ 
\hline
\end{tabular}
  }
\label{tab:fewshots}
\vskip  -.5 cm
\end{table}

\begin{table}[h]
\vskip -.1cm
\centering
\caption{\small Ablation Study of SQLPrompt (without Mix LLMs) 
}\label{tab:sqlprompt-ablation}
\vskip -2mm
\label{tab:ablation-nomix}
\resizebox{0.8\columnwidth}{!}{%
\begin{tabular}{lll}
\hline
{\color[HTML]{000000} \textbf{}}                         & {\color[HTML]{000000} \textbf{EX}} & {\color[HTML]{000000} \textbf{TS}} \\ \hline
{\color[HTML]{000000} \begin{tabular}[c]{@{}l@{}}\textbf{SQLPrompt} (Prompt Design \\ + Consistency \\ + Execution Filtering\\ +MixPrompt)\end{tabular}} &
  {\color[HTML]{000000} 70.5} &
  {\color[HTML]{000000} 63.2}  \\ \hline
\multicolumn{1}{l}{{\color[HTML]{000000} No MixPrompt}} & {\color[HTML]{000000} 67.7}        & {\color[HTML]{000000} 61.3 ({\color{red} $\downarrow 1.9$}) }        \\\hline
{\color[HTML]{000000} \begin{tabular}[c]{@{}l@{}}Only Schema (No primary, \\ No foreignkeys, no DB content)\end{tabular}} &
  {\color[HTML]{000000} 66.4} &
  {\color[HTML]{000000} 57.3} ({\color{red} $\downarrow 5.9$}) \\ \hline
{\color[HTML]{000000} No Consistency}                    & {\color[HTML]{000000} 55.9}        & {\color[HTML]{000000} 49.6 ({\color{red} $\downarrow 13.6$}) }        \\ \hline
{\color[HTML]{000000} No  Execution Filtering}           & {\color[HTML]{000000} 55.2}        & {\color[HTML]{000000} 48.7 ({\color{red} $\downarrow 14.5$}) }        \\ \hline
\end{tabular}%
}
\vskip   -0.3cm
\end{table}

\begin{table}[h]
\centering
\caption{\small Ablation Study: SQLprompt with Mix LLMs}\label{tab:mixture}
\vskip  -0.2cm
\resizebox{0.7\columnwidth}{!}{%
\begin{tabular}{lcccc}
\hline
\multirow{2}{*}{Num of Mixture}  &      \multicolumn{2}{c}{\text{Zero-shots}} & \multicolumn{2}{c}{\text{Few-shots}} \\  \cline{2-3} \cline{4-5} 
 
 & 2 & 4 & 6 & 16 \\
\hline
EX & 74 & 76.6& 77.3 & 77.1  \\ 
TS & 65.5 & 68.0 & 68.3 & 68.6\\ 
\hline
\end{tabular}
}
\vskip  -.5cm
\end{table}
\section{Experiments}
\paragraph{Tasks and datasets:}
We consider the cross-domain large-scale Text-to-SQL benchmark, Spider \citep{yu2018spider} that contains 7000 training samples across 166 databases and 1034 evaluation samples (`Dev split') across 20 databases. 
\paragraph{Models:}
\textbf{PaLM FLAN 540B} is a PaLM model variant \citep{chowdhery2022palm} with
540 billion parameters fine-tuned on a collection of tasks phrased as instructions. FLAN \citep{chung2022scaling} is a reference to the way of fine-tuning that reflects instructions being given in the prompt. \textbf{PaLM-62B} is a PaLM variant with 62 billion parameters trained on 1.3T tokens following the \cite{hoffmann2022training} 
\textbf{PaLM FLAN 62B} is FLAN fine-tuned variant.
\textbf{Quantization} is applied to above models with \emph{qntz}, that reduces the precision albeit increased inference efficiency. 

\textbf{Fine-tuning baselines}: 
\textbf{PICARD} \citep{scholak2021picard} employs incremental parsing to constrain auto-regressive decoding.
\textbf{RASAT} \cite{qi2022rasat} is a transformer model that integrates relation-aware self-attention and constrained auto-regressive decoders. 
\textbf{RESDSQL} \cite{li2023decoupling} decouples schema linking and skeleton parsing using a ranking-enhanced encoding and skeleton-aware decoding framework. 

\textbf{In-context learning baselines}:  
\citep{rajkumar2022evaluating} comprehensively evaluate the Text-to-SQL ability of CodeX and GPT3, while \citep{liu2023comprehensive} conduct a comprehensive evaluation on ChatGPT.

 \paragraph{Evaluation:}

We consider two commonly-used evaluation metrics: execution accuracy (EX) and test-suite accuracy (TS) \citep{zhong2020semantic}, where EX measures if SQL execution outcome matches ground truth. TS assesses each query by running \textbf{multiple tests} against randomly generated database with same schema (EX only evaluates on one test). So TS reduces false positives from EX and hence can be more precise. 
Exact match evaluation is not performed, as multiple correct SQLs exist for one query. 
\vskip -4mm
\section{Results}
\vskip -1mm
Table-\ref{tab:sota} 
presents the comparison between SQLPrompt and the previous methods for in-context learning and fine-tuning. For in-context learning, SQLPrompt outperforms ChatGPT (with their recommended prompts) by a large margin: $\uparrow 7\%$ for execution accuracy (EX) and $\uparrow 8.1\%$ for test suite accuracy (TS). Examples of SQL generated by SQLPrompt are provided in Table~\ref{tab:LLMgeneration} in Appendix. 
\paragraph{Ablation study}SQLPrompt consists of multiple components: prompt design, execution-based consistency decoding, MixPrompt, and MixLLMs. 
The effect of
MixPrompt prompt is shown in Table~\ref{tab:fewshots}, leading to
$2 \%+$ improvement over single prompt. The effect of mixing LLMs
is shown in Table~\ref{tab:mixture}, leading to additional $3 \%$ over single LLM in test suite accuracy (details description of mixLLMs are provided in Appendix \ref{mixllm}). 

To shed light into the impact of these components,
ablation studies are presented in Table~\ref{tab:sqlprompt-ablation}. All listed components are observed to be useful, yielding $2 \sim 15  \%$
improvements.
Notably, without consistency decoding and keeping all others unchanged, the performance decreases by $13.6\%$; that is it contributes by $13.6\%$;  Without "execution error filtering" and keeping other components unchanged, the results decrease by $14.5\%$. "MixPrompt" contributes by $2\%$ and adding "Primary Keys/Foreign Keys/DB content" contributes by $5.9\%$.

\clearpage


\section*{Limitations}
The limitation of this work is that the method queries multiple prompt designs and LLMs, that can be expensive and time consuming. Although combining multiple prompt designs and LLMs are promising to improve performance, future work can explore more efficient and effective ways to combine them to save cost.  

\bibliography{anthology,custom}
\bibliographystyle{acl_natbib}
\clearpage
\appendix
\section{Text-to-SQL challenges and prompt design with primary/foreign keys and database content} \label{sec:promptdesign-explained}
\begin{figure}[ht]
\includegraphics[width=0.45\textwidth]{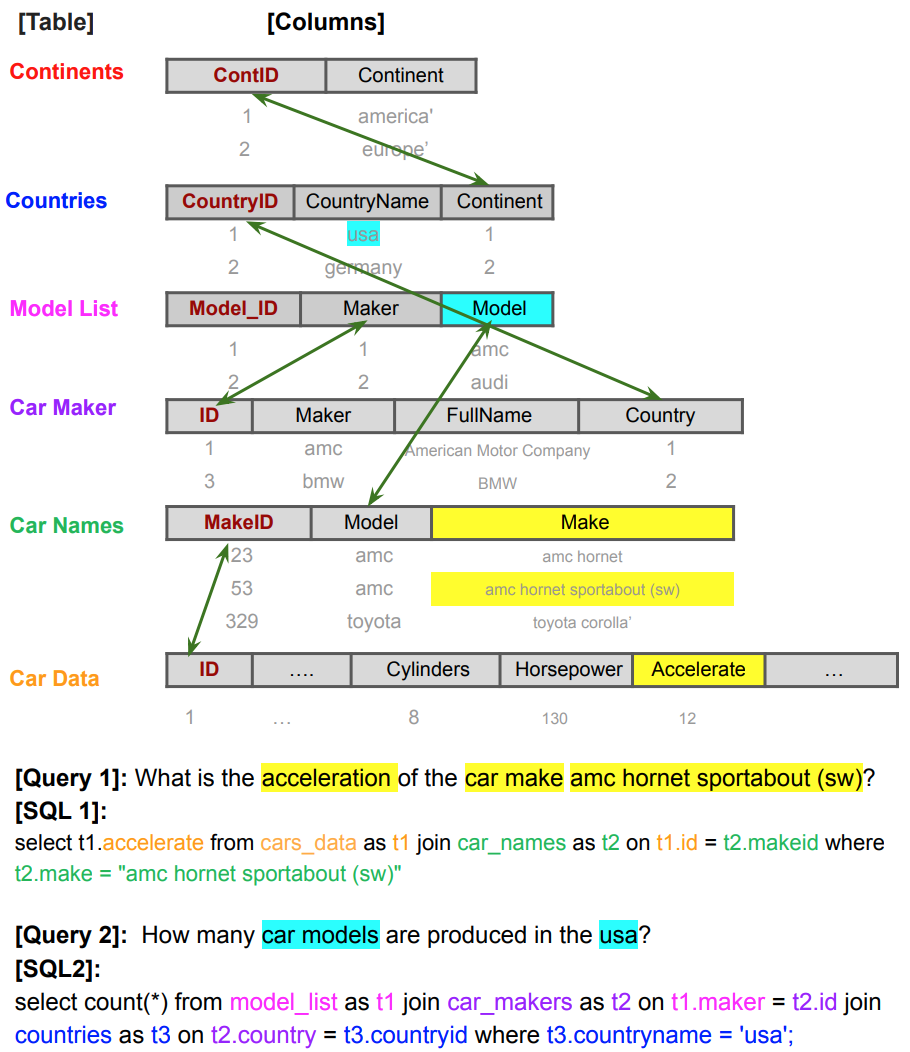}
\caption{One database schema with two query questions and true SQL as demo.  {\color{carmine} Dark red} are primary keys. {\color{forestgreen(traditional)} Dark green} arrows are foreign keys joining different tables. {\color{gray} Light gray} is the context (values) in database (or table). Both primary key and foreign keys are given in the database schema. The highlighted (yellow or cyan) are the part of schema that are used to solve Query 1 and 2 respectively. Colors are simply for easy visualization. Same color, same table. 
\label{fig:overivew}}
\end{figure}

\begin{figure}[ht]
\includegraphics[width=0.45\textwidth]{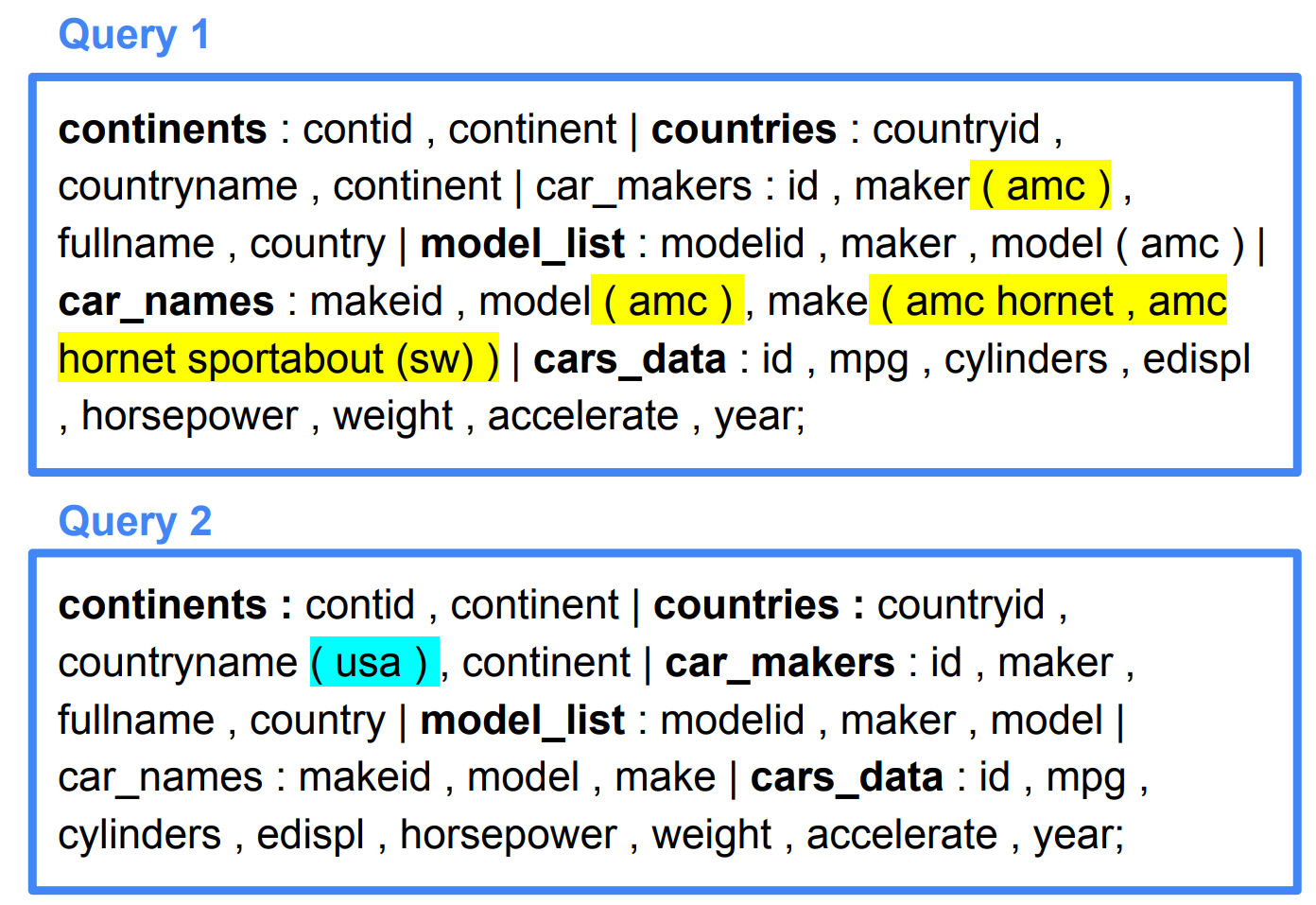}
\caption{\textbf{Example of database with content}: examples in Fig~\ref{fig:overivew}. Highlighted are database content for different queries. Following previous work \citep{xie2022unifiedskg, scholak2021picard}, only the relevant database content values are included. So different query questions have different database content value. \label{fig:db-content}}
\vskip -0.5cm
\end{figure}

Fig.~\ref{fig:overivew} shows a Text-to-SQL example from Spider. Fig.~\ref{fig:overivew} demonstrates the necessity of including primary and foreign keys, and content of database. The data schema contains multiple tables. Each table has multiple columns. \textbf{Primary keys} are the columns that uniquely identify a row in a table. Primary keys are important, as some columns might specifically be challenging and it might be beneficial to include them specifically as prompts, such as in Query 1 of Fig.~\ref{fig:overivew} where "t2.makeid" may be mistakenly written as "t2.id" without proper emphasis. \textbf{Foreign key} is a column or combination of columns that is used to establish and enforce a link between the data in two tables. For example, in Fig~\ref{fig:overivew} , Column Maker of Table Model list is equivalent to Column ID of Car Maker. By including foreign keys into prompt, LLMs can know how to join different tables. Otherwise, it can be ambiguous to link multiple tables, especially for complex data schema or schema with confusing column names. For example, Column Maker in Table Model list is not the same as Column Maker in Table Car Maker. Although they both called column "Maker", one is number and the other is string. Instead due to foreign keys, we known Column Maker of Table Model List is equivalent to Column ID in Table Car maker.   Additionally, including relevant \textbf{database content value}, as seen in \citep{xie2022unifiedskg, scholak2021picard}, is necessary as they help identify which columns are relevant to key words in the query question, such as in Fig.~\ref{fig:overivew}, Query1's key information is "amc honrnet sportabout (sw)", however, without adding database content value, we do not know which columns contain the value of the key information. e.g. is it Column Maker of Table Model List? Is it Column Maker of Table Car Maker? or Is it Column Make of Table Car Names? Only by including database content values, LLM can know it should use The column of Make of Table Car Names. 
Note that the database content values are questions depended. Only content values that are related with questions is included into prompt. See Fig~\ref{fig:db-content}. Note not all the content values are included. So there is not problem if the number of database contents is very large. As for how to extract relevant database content values regarding the query questions, we follow \citep{xie2022unifiedskg, scholak2021picard}, where all the content values are compared against the query questions, and only top few ones that match the query question the best are included. 

\section{Prompt design examples}
We show the prompt design for a example in Spider dataset. 

\subsection{Concise prompt design}\label{sec:concise}

"This is a task converting text into SQL statement. We will first given the dataset schema and then ask a question in text. You are asked to generate SQL statement.
 Here is the test question to be anwered: Convert text to SQL:
 \textbf{[Schema (values)]}: | car\_1 | continents : contid , continent | countries : countryid , countryname , continent | car\_makers : id , maker ( amc ) , fullname , country | model\_list : modelid , maker , model ( amc ) | car\_names : makeid , model ( amc ) , make ( amc hornet , amc hornet sportabout (sw) ) | cars\_data : id , mpg , cylinders , edispl , horsepower , weight , accelerate , year;
 \textbf{[Column names (type)]}: continents : contid (number) | continents : continent (text) | countries : countryid (number) | countries : countryname (text) | countries : continent (number) | car\_makers : id (number) | car\_makers : maker (text) | car\_makers : fullname (text) | car\_makers : country (text) | model\_list : modelid (number) | model\_list : maker (number) | model\_list : model (text) | car\_names : makeid (number) | car\_names : model (text) | car\_names : make (text) | cars\_data : id (number) | cars\_data : mpg (text) | cars\_data : cylinders (number) | cars\_data : edispl (number) | cars\_data : horsepower (text) | cars\_data : weight (number) | cars\_data : accelerate (number) | cars\_data : year (number);
 \textbf{[Primary Keys]}: continents : contid | countries : countryid | car\_makers : id | model\_list : modelid | car\_names : makeid | cars\_data : id;
 \textbf{[Foreign Keys]}: countries : continent equals continents : contid | car\_makers : country equals countries : countryid | model\_list : maker equals car\_makers : id | car\_names : model equals model\_list : model | cars\_data : id equals car\_names : makeid
 \textbf{[Q]}: What is the accelerate of the car make amc hornet sportabout (sw)?;
 \textbf{[SQL]}: "
 
\subsection{Verbose prompt design}\label{sec:verbose}
"This is a task converting text into SQL statement. We will first given the dataset schema and then ask a question in text. You are asked to generate SQL statement.
 Here is the test question to be anwered: Let us take a question and turn it into a SQL statement about database tables. There are 6 tables. Their titles are: continents, countries, car\_makers, model\_list, car\_names, cars\_data. Table 1 is continents, and its column names and types are: ContId (Type is number), Continent (Type is text). Table 2 is countries, and its column names and types are: CountryId (Type is number), CountryName (Type is text), Continent (Type is number). Table 3 is car\_makers, and its column names and types are: Id (Type is number), Maker (Type is text), FullName (Type is text), Country (Type is text). Table 4 is model\_list, and its column names and types are: ModelId (Type is number), Maker (Type is number), Model (Type is text). Table 5 is car\_names, and its column names and types are: MakeId (Type is number), Model (Type is text), Make (Type is text). Table 6 is cars\_data, and its column names and types are: Id (Type is number), MPG (Type is text), Cylinders (Type is number), Edispl (Type is number), Horsepower (Type is text), Weight (Type is number), Accelerate (Type is number), Year (Type is number). The primary keys are: contid from Table continents, countryid from Table countries, id from Table car\_makers, modelid from Table model\_list, makeid from Table car\_names, id from Table cars\_data. The foreign keys are: continent from Table countries is equivalent with contid from Table continents, country from Table car\_makers is equivalent with countryid from Table countries, maker from Table model\_list is equivalent with id from Table car\_makers, model from Table car\_names is equivalent with model from Table model\_list, id from Table cars\_data is equivalent with makeid from Table car\_names. Use foreign keys to join Tables. Columns with relevant values: Table car\_makers Column maker have values: amc; Table model\_list Column model have values: amc; Table car\_names Column model have values: amc; Table car\_names Column make have values: amc hornet, amc hornet sportabout (sw);  Only use columns with relevant values to generate SQL.  Let us take a text question and turn it into a SQL statement about database tables. The question is: What is the accelerate of the car make amc hornet sportabout (sw)? The corresponding SQL is: "
 \vskip -4cm
\section{ChatGPT's default prompt design} \label{sec:openai-prompt}
"Complete sqlite SQL query only and with no explanation Sqlite SQL tables, with their properties: continents(ContId, Continent); countries(CountryId, CountryName, Continent); car\_makers(Id,Maker, FullName, Country); mode\_list(ModelId, Maker, Model); car\_names(MakeId, Model, Make); car\_data(Id, MPG, Cylinders, Edispl, Horsepower, Weight, Accelerate, Year).  What is the accelerate of the car make amc hornet sportabout (sw)? SELECT"
\begin{table}[ht]
\centering
\caption{Additional few-shot learning results.}
\vskip -3mm
\resizebox{0.5\textwidth}{!}{
\begin{tabular}{lcccccc}
\hline

\multirow{2}{*}{Methods/Datasets} & 
\multicolumn{2}{c}{\textbf{Concise}} & \multicolumn{2}{c}{\textbf{Verbose}} & \multicolumn{2}{c}{\textbf{MixPrompt}}  \\ 
 & EX & TS & EX & TS & EX & TS \\ 
\hline
PaLM FLAN 62B q   & 68.1  &61.7 &71.9  & 64.3 &   74.5  &  66.7      \\
\hline
\end{tabular}
}
\label{tab:fewshots-1}
\end{table}

\begin{table*}[h]
\scriptsize
\centering
\resizebox{2\columnwidth}{!}{%
\begin{tabular}{p{3cm}|p{15cm}} 
\hline
Question & What are the names of the singers whose birth years are either 1948 or 1949? \\
Ground Truth & SELECT Name FROM singer WHERE Birth\_Year  =  1948 OR Birth\_Year  =  1949 \\
LLM Generation & SELECT name FROM singer WHERE birth\_year = 1948 OR birth\_year = 1949 \\
Execution Accuracy & 1 \\  \hline
Question & What is the name of the singer with the largest net worth? \\
Ground Truth & SELECT Name FROM singer ORDER BY Net\_Worth\_Millions DESC LIMIT 1 \\
LLM Generation & SELECT T1.name FROM singer AS T1 ORDER BY T1.net\_worth\_millions Desc LIMIT 1 \\
Execution Accuracy & 1 \\  \hline
Question & What is the name of the singer who is worth the most? \\
Ground Truth & SELECT Name FROM singer ORDER BY Net\_Worth\_Millions DESC LIMIT 1 \\
LLM Generation & SELECT T1.name FROM singer AS T1 JOIN song AS T2 ON T1.singer\_id  =  T2.singer\_id ORDER BY T1.net\_worth\_millions Desc LIMIT 1 \\
Execution Accuracy & 1 \\  \hline
Question & Show different citizenship of singers and the number of singers of each citizenship. \\
Ground Truth & SELECT Citizenship ,  COUNT(*) FROM singer GROUP BY Citizenship \\
LLM Generation & SELECT T1.citizenship ,  count(*) FROM singer AS T1 GROUP BY T1.citizenship \\
Execution Accuracy & 1 \\  \hline
Question & For each citizenship, how many singers are from that country? \\
Ground Truth & SELECT Citizenship ,  COUNT(*) FROM singer GROUP BY Citizenship \\
LLM Generation & SELECT Citizenship ,  count(*) FROM singer GROUP BY Citizenship \\
Execution Accuracy & 1 \\  \hline
Question & Please show the most common citizenship of singers. \\
Ground Truth & SELECT Citizenship FROM singer GROUP BY Citizenship ORDER BY COUNT(*) DESC LIMIT 1 \\
LLM Generation & SELECT citizenship FROM singer GROUP BY citizenship ORDER BY count(*) DESC LIMIT 1 \\
Execution Accuracy & 1 \\  \hline
Question & What is the most common singer citizenship ? \\
Ground Truth & select citizenship from singer group by citizenship order by count(*) desc limit 1 \\
LLM Generation & SELECT citizenship FROM singer GROUP BY citizenship ORDER BY count(*) DESC LIMIT 1 \\
Execution Accuracy & 1 \\  \hline
Question & Show different citizenships and the maximum net worth of singers of each citizenship. \\
Ground Truth & SELECT Citizenship ,  max(Net\_Worth\_Millions) FROM singer GROUP BY Citizenship \\
LLM Generation & SELECT T1.citizenship ,  max(T1.net\_worth\_millions) FROM singer AS T1 GROUP BY T1.citizenship \\
Execution Accuracy & 1 \\  \hline
Question & For each citizenship, what is the maximum net worth? \\
Ground Truth & SELECT Citizenship ,  max(Net\_Worth\_Millions) FROM singer GROUP BY Citizenship \\
LLM Generation & SELECT T1.citizenship ,  max(T1.net\_worth\_millions) FROM singer AS T1 GROUP BY T1.citizenship \\
Execution Accuracy & 1 \\  \hline
Question & Show titles of songs and names of singers. \\
Ground Truth & SELECT T2.Title ,  T1.Name FROM singer AS T1 JOIN song AS T2 ON T1.Singer\_ID  =  T2.Singer\_ID \\
LLM Generation & SELECT T1.title ,  T2.name FROM song AS T1 JOIN singer AS T2 ON T1.singer\_id = T2.singer\_id \\
Execution Accuracy & 1 \\  \hline
Question & What are the song titles and singer names? \\
Ground Truth & SELECT T2.Title ,  T1.Name FROM singer AS T1 JOIN song AS T2 ON T1.Singer\_ID  =  T2.Singer\_ID \\
LLM Generation & SELECT T1.title ,  T2.name FROM song AS T1 JOIN singer AS T2 ON T1.singer\_id = T2.singer\_id \\
Execution Accuracy & 1 \\  \hline
Question & Show distinct names of singers that have songs with sales more than 300000. \\
Ground Truth & SELECT DISTINCT T1.Name FROM singer AS T1 JOIN song AS T2 ON T1.Singer\_ID  =  T2.Singer\_ID WHERE T2.Sales  >  300000 \\
LLM Generation & SELECT DISTINCT T1.name FROM singer AS T1 JOIN song AS T2 ON T1.singer\_id = T2.singer\_id WHERE T2.sales  >  300000 \\
Execution Accuracy & 1 \\  \hline
Question & what are the different names of the singers that have sales more than 300000? \\
Ground Truth & SELECT DISTINCT T1.Name FROM singer AS T1 JOIN song AS T2 ON T1.Singer\_ID  =  T2.Singer\_ID WHERE T2.Sales  >  300000 \\
LLM Generation & SELECT T1.name FROM singer AS T1 JOIN song AS T2 ON T1.singer\_id = T2.singer\_id WHERE T2.sales  >  300000 \\
Execution Accuracy & 1 \\  \hline
Question & Show the names of singers that have more than one song. \\
Ground Truth & SELECT T1.Name FROM singer AS T1 JOIN song AS T2 ON T1.Singer\_ID  =  T2.Singer\_ID GROUP BY T1.Name HAVING COUNT(*)  >  1 \\
LLM Generation & SELECT T1.name FROM singer AS T1 JOIN song AS T2 ON T1.singer\_id = T2.singer\_id GROUP BY T1.singer\_id HAVING COUNT(*)  >  1 \\
Execution Accuracy & 1 \\  \hline
Question & What are the names of the singers that have more than one songs? \\
Ground Truth & SELECT T1.Name FROM singer AS T1 JOIN song AS T2 ON T1.Singer\_ID  =  T2.Singer\_ID GROUP BY T1.Name HAVING COUNT(*)  >  1 \\
LLM Generation & SELECT T1.name FROM singer AS T1 JOIN song AS T2 ON T1.singer\_id = T2.singer\_id GROUP BY T1.singer\_id HAVING COUNT(*)  >  1 \\
Execution Accuracy & 1 \\  \hline
Question & Show the names of singers and the total sales of their songs. \\
Ground Truth & SELECT T1.Name ,  sum(T2.Sales) FROM singer AS T1 JOIN song AS T2 ON T1.Singer\_ID  =  T2.Singer\_ID GROUP BY T1.Name \\
LLM Generation & SELECT T1.name ,  sum(T2.sales) FROM singer AS T1 JOIN song AS T2 ON T1.singer\_id = T2.singer\_id GROUP BY T1.name \\
Execution Accuracy & 1 \\  \hline
Question & For each singer name, what is the total sales for their songs? \\
Ground Truth & SELECT T1.Name ,  sum(T2.Sales) FROM singer AS T1 JOIN song AS T2 ON T1.Singer\_ID  =  T2.Singer\_ID GROUP BY T1.Name \\
LLM Generation & SELECT T1.name ,  sum(T2.sales) FROM singer AS T1 JOIN song AS T2 ON T1.singer\_id  =  T2.singer\_id GROUP BY T1.name \\
Execution Accuracy & 1 \\  \hline
Question & List the name of singers that do not have any song. \\
Ground Truth & SELECT Name FROM singer WHERE Singer\_ID NOT IN (SELECT Singer\_ID FROM song) \\
LLM Generation & SELECT name FROM singer WHERE singer\_id NOT IN ( SELECT singer\_id FROM song ) \\
Execution Accuracy & 1 \\  \hline
Question & What is the sname of every sing that does not have any song? \\
Ground Truth & SELECT Name FROM singer WHERE Singer\_ID NOT IN (SELECT Singer\_ID FROM song) \\
LLM Generation & SELECT T1.name FROM singer AS T1 JOIN song AS T2 ON T1.singer\_id = T2.singer\_id WHERE T2.singer\_id IS NULL \\
Execution Accuracy & 0 \\  \hline
Question & Show the citizenship shared by singers with birth year before 1945 and after 1955. \\
Ground Truth & SELECT Citizenship FROM singer WHERE Birth\_Year  <  1945 INTERSECT SELECT Citizenship FROM singer WHERE Birth\_Year  >  1955 \\
LLM Generation & SELECT T1.citizenship FROM singer AS T1 JOIN song AS T2 ON T1.singer\_id  =  T2.singer\_id WHERE T1.birth\_year  <  1945 INTERSECT SELECT T1.citizenship FROM singer AS T1 JOIN song AS T2 ON T1.singer\_id  =  T2.singer\_id WHERE T1.birth\_year  >  1955 \\
Execution Accuracy & 0 \\  \hline
Question & What are the citizenships that are shared by singers with a birth year before 1945 and after 1955? \\
Ground Truth & SELECT Citizenship FROM singer WHERE Birth\_Year  <  1945 INTERSECT SELECT Citizenship FROM singer WHERE Birth\_Year  >  1955 \\
LLM Generation & SELECT citizenship FROM singer WHERE birth\_year  <  1945 INTERSECT SELECT citizenship FROM singer WHERE birth\_year  >  1955 \\
Execution Accuracy & 1 \\  \hline
Question & How many available features are there in total? \\
Ground Truth & SELECT count(*) FROM Other\_Available\_Features \\
LLM Generation & SELECT count(*) FROM other\_available\_features \\
Execution Accuracy & 1 \\  \hline
Question & What is the feature type name of feature AirCon? \\
Ground Truth & SELECT T2.feature\_type\_name FROM Other\_Available\_Features AS T1 JOIN Ref\_Feature\_Types AS T2 ON T1.feature\_type\_code  =  T2.feature\_type\_code WHERE T1.feature\_name  =  "AirCon" \\
LLM Generation & SELECT T1.feature\_type\_name FROM ref\_feature\_types AS T1 JOIN other\_available\_features AS T2 ON T1.feature\_type\_code = T2.feature\_type\_code WHERE T2.feature\_name = "AirCon" \\
Execution Accuracy & 1 \\  \hline
Question & Show the property type descriptions of properties belonging to that code. \\
Ground Truth & SELECT T2.property\_type\_description FROM Properties AS T1 JOIN Ref\_Property\_Types AS T2 ON T1.property\_type\_code  =  T2.property\_type\_code GROUP BY T1.property\_type\_code \\
LLM Generation & SELECT property\_type\_description FROM ref\_property\_types WHERE property\_type\_code IN (SELECT property\_type\_code FROM properties) \\
Execution Accuracy & 1 \\  \hline
Question & What are the names of properties that are either houses or apartments with more than 1 room? \\
Ground Truth & SELECT property\_name FROM Properties WHERE property\_type\_code  =  "House" UNION SELECT property\_name FROM Properties WHERE property\_type\_code  =  "Apartment" AND room\_count  >  1 \\
LLM Generation & SELECT property\_name FROM properties WHERE property\_type\_code  =  'House' OR property\_type\_code  =  'Apartment' AND room\_count  >  1 \\
Execution Accuracy & 1 \\  \hline
\end{tabular}
}
\caption{SQLPrompt: randomly sampled generated SQL from LLM}
\label{tab:LLMgeneration}
\end{table*}

\section*{SQLPrompt output} 
We show a list of real output generated by $SQLPrompt$. See Table~\ref{tab:LLMgeneration}

\section{Selection of few-shot examples}
We also provide results in few-shot learning setting with a different set of examples. Compared to Table~\ref{tab:fewshots} (e.g. 66.6), the two results are very close, indicating changing the few-shot examples doesn't improve performance when using the same prompt design. 




\section{Experiment details}
Querying LLMs might not require TPU or GPU -- a CPU would suffice. We use the number of consistency repeats of 32 and the sampling temperature of 0.5. 

\section{MixLLMs implementation}\label{mixllm}
The effect of MixLLMs of \emph{SQLPrompt} shows in Table \ref{tab:mixture}. When the number of mixture is less than 4, we use zero-shot results from Table~\ref{tab:zeroshots}. 
For example, with 4  mixtures, we combine all the 4 models in Table~\ref{tab:zeroshots}: PaLM FLAN 62B q: Concise or Verbose prompt design; PaLM FLAN 540B q: Concise or Verbose. When number of mixture is greater than 4, we include few-shot results. Note that most of the components in SQLPrompt can be applied to other in-context learning methods. 
We consider LLMs with different few-shot examples as a ``different models'', as few-shot demonstrations can influence the behaviors of LLMs. 

\end{document}